\documentclass[letterpaper, 10 pt, conference]{ieeeconf}  

\IEEEoverridecommandlockouts                              

\usepackage{amsmath} 
\usepackage{amssymb}  
\usepackage{booktabs}
\usepackage{multirow}
\usepackage{graphicx}
\usepackage{color}   
\usepackage[normalem]{ulem}
\usepackage{xcolor}
\usepackage{pifont}

\title{\LARGE \bf
Keyframe-Based Feed-Forward Visual Odometry
}

\author{
Weichen Dai$^{1}$ Wenhan Su$^{1}$, Da Kong$^{2}$, Yuhang Ming$^{1}$, and Wanzeng Kong$^{1}$
\thanks{$^{1}$Key Laboratory of Brain Machine Collaborative Intelligence of Zhejiang Province, School of Computer Science, Hangzhou Dianzi University, Hangzhou, China}%
\thanks{$^{2}$Technion Autonomous Systems Program, Technion - Israel Institute of Technology, Haifa, Israel }%
}

\begin{document}

\maketitle
\thispagestyle{empty}
\pagestyle{empty}

\begin{abstract}

The emergence of visual foundation models has revolutionized visual odometry~(VO) and SLAM, enabling pose estimation and dense reconstruction within a single feed-forward network. However, unlike traditional pipelines that leverage keyframe methods to enhance efficiency and accuracy, current foundation model based methods, such as VGGT-Long, typically process raw image sequences indiscriminately. This leads to computational redundancy and degraded performance caused by low inter-frame parallax, which provides limited contextual stereo information. Integrating traditional geometric heuristics into these methods is non-trivial, as their performance depends on high-dimensional latent representations rather than explicit geometric metrics. To bridge this gap, we propose a novel keyframe-based feed-forward VO. Instead of relying on hand-crafted rules, our approach employs reinforcement learning to derive an adaptive keyframe policy in a data-driven manner, aligning selection with the intrinsic characteristics of the underlying foundation model. We train our agent on TartanAir dataset and conduct extensive evaluations across several real-world datasets. Experimental results demonstrate that the proposed method achieves consistent and substantial improvements over state-of-the-art feed-forward VO methods.

\end{abstract}


\section{INTRODUCTION}

Visual Odometry (VO) and Simultaneous Localization and Mapping (SLAM) constitute the backbone for robotics, enabling autonomous systems to estimate their pose and reconstruct maps of unknown environments using visual input. Traditional VO and SLAM approaches~\cite{engel2017direct,campos2021orb,teed2024deep}, rely on multi-stage pipelines~\cite{cadena2016past} that involve feature extraction, tracking and nonlinear optimization under explicit geometric constraints~\cite{barfoot2024state}. However, these methods suffer from inherent limitations: they require extensive expert knowledge to carefully tune parameters and design heuristics in order to handle a wide range of corner cases encountered in real-world scenarios, including dynamic scenes~\cite{dai2020rgb}, low-texture environments~\cite{dai2019multi}, and severe illumination changes~\cite{dai2021multi}.

The recent emergence of feed-forward visual foundation models has made it possible to establish a unified VO pipeline that requires no calibration and tuning with great generalization capabilities. Pioneering works such as DUSt3R~\cite{wang2024dust3r} and its successor MASt3R~\cite{leroy2024grounding} have demonstrated the feasibility of jointly estimating camera parameters and dense 3D geometry from uncalibrated image pairs. Subsequent models, including CUT3R~\cite{wang2025continuous} and Fast3R~\cite{yang2025fast3r}, have further optimized this paradigm. Most recently, VGGT~\cite{wang2025vggt} achieved state-of-the-art reconstruction quality, generating stable and accurate local 3D maps directly from raw monocular RGB input. Extensions such as VGGT-SLAM~\cite{maggio2025vggt} and VGGT-Long~\cite{deng2025vggt} have introduced chunking and loop closure to achieve globally aligned long-term pose estimation. Despite these advances, these methods generally lack filtering for input frames. They do not mitigate the negative impact of low-quality and redundant frames on local estimation and do not address the computational and memory reduction caused by similar input frames.

Extensive research in both VO/SLAM~\cite{younes2017keyframe} and Structure-from-Motion (SfM)~\cite{conti2024range} has demonstrated that a well-designed keyframe method can effectively improve performance while reducing computational burden. However, translating traditional keyframe method to foundation models is not easy. Conventional methods rely on explicit geometric metrics, whereas foundation models like VGGT operate as "black boxes" with complex, high-dimensional latent spaces. Designing a rule-based keyframe method that aligns with the internal characteristics of these models, without explicit knowledge of their internal mechanics, remains a significant challenge.

To this end, we propose a keyframe-based feed-forward VO method that using VGGT as the backbone with a reinforcement learning–based keyframe method. The keyframe method adaptively learns, in a data-driven manner, to align with the intrinsic characteristics of VGGT. This enables more efficient management of the sliding-window inputs and leads to improved overall localization accuracy.

We train the proposed method on the synthetic TartanAir dataset~\cite{wang2020tartanair} and conduct extensive experiments on several widely used real-world datasets, including EuRoC~\cite{burri2016euroc}, TUM-RGBD~\cite{sturm2012benchmark}, and KITTI~\cite{geiger2013vision}. The results consistently demonstrate significant performance improvements.

Our main contributions can be summarized as follows:
\begin{itemize}
    \item We introduced a keyframe-based feed-forward VO method that leveraging the ability of VGGT with a data-driven RL based keyframe method. By aligning input frame selection with the characteristics of VGGT.
    \item We introduced an adaptive keyframe method into 3D vision foundation models like VGGT. By aligning input frame selection with the intrinsic representation characteristics of foundation models, the proposed approach significantly improves localization performance.
    \item The experiment results demonstrate strong generalization capability of the proposed method cross real-world datasets.
\end{itemize}

\section{RELATED WORK}

\subsection{The State of VO}
In the field of VO, two primary technical approaches dominate: classic methods and end-to-end learning methods.

Classic methods have demonstrated exceptional accuracy and stability in industrial-grade applications. The tracking methods are categorized into three typical paradigms: direct visual odometry, feature point matching, and hybrid semi-direct methods. Direct methods, exemplified by DSO~\cite{engel2017direct}, use pixel intensity to identify key image regions and analyze their motion across frames to estimate camera pose. Feature point methods like ORB-SLAM3~\cite{campos2021orb}, extract feature points from images and track their positional changes to estimate motion. Semi-direct methods such as SVO~\cite{forster2016svo} combine feature point detection with brightness analysis of surrounding image patches to estimate motion.

The recent paradigm shift driven by deep learning has given rise to data-driven, end-to-end methods~\cite{wang2017deepvo}~\cite{wang2021tartanvo}. In representative work, DPVO~\cite{teed2024deep} employs differentiable optimization layers and recurrent network architectures to realize a motion estimation system based on image patch tracking. LEAP-VO~\cite{chen2024leap} employs long-term point tracking with anchor-based dynamic estimation and temporal probability modeling to build a robust visual odometry system for dynamic scenes. MAC-VO~\cite{qiu2025mac} introduces a stereo visual odometry framework that leverages a metrics-aware covariance model for keypoint selection and residual weighting in pose graph optimization.

\subsection{3D Vision Foundation Models}
In recent, end-to-end foundation models have emerged as one of the most prominent research frontiers in the field of 3D computer vision. Primarily built upon the transformer architecture~\cite{vaswani2017attention}, these models aim to achieve dense geometric reconstruction directly from images without requiring prior knowledge of camera poses or calibration parameters.

Pioneering research, represented by DUSt3R~\cite{wang2024dust3r} and its successor MASt3R~\cite{leroy2024grounding}, first demonstrated the feasibility of jointly estimating camera parameters and dense 3D structures from uncalibrated image pairs. Subsequently, models such as CUT3R~\cite{wang2025continuous} and Fast3R~\cite{yang2025fast3r} have further refined this technical paradigm, achieving a superior balance between computational efficiency and reconstruction accuracy. 

Currently, the state-of-the-art model VGGT~\cite{wang2025vggt} is capable of generating highly stable and precise local 3D maps directly from raw RGB inputs. Concurrently, new novel models like ${\pi}^3$~\cite{wang2025pi}, Depth Anything 3~\cite{lin2025depth}, and MapAnything~\cite{keetha2025mapanything} are continue to drive rapid advancements in this domain.

However, a significant limitation shared by these models is their substantial computational and memory overhead, which severely restricts their practical deployment in tasks such as long-sequence visual localization or large-scale scene reconstruction. To address this challenge, recent works including VGGT-SLAM~\cite{maggio2025vggt}, VGGT-Long~\cite{deng2025vggt}, and FlashVGGT~\cite{wang2025flashvggt} have incorporated sliding-window mechanisms with loop closure, enabling VGGT to efficiently process long image sequences.

\subsection{Keyframe Management Mechanisms}

There are three categories of keyframe management in VO methods: heuristic methods, probability-based methods, and learning-based methods.

Heuristic methods dominate the realm of keyframe management~\cite{dias2023keyframe}. They typically rely on predefined rules based on distance or visual appearance to select keyframe. In ORB-SLAM3~\cite{campos2021orb}, keyframe insertion is related to viewpoint variation, tracking quality, and redundancy control. DSO~\cite{engel2017direct} selects keyframe by considering field-of-view changes, occlusion/disocclusion phenomena, and camera exposure time variations. SVO~\cite{forster2016svo} determines new keyframe by comparing the Euclidean distance to existing keyframe with the average scene depth. Chen et al.~\cite{chen2021dynamic} introduced a proportional–derivative(PD) controller to adaptively adjust the threshold for keyframe selection.

Probability-based methods, in contrast to heuristic approaches, are more theoretically grounded but less frequently applied. The majority of these methods make decisions probabilistically, based on the geometric information of frames. Knoblauch et al.~\cite{knoblauch2011non} use relGRIC to detect degeneracy and assess keyframe based on point quality and scene coverage. Pumarola et al.~\cite{pumarola2017pl} select keyframe by comparing the entropy ratio between frames to a predefined threshold. Schmuck et al.~\cite{schmuck2019redundancy} quantify keyframe information using Mutual Information, assessing content via conditional Mutual Information.

Although interest in learning-based keyframe management is growing, the field remains relatively nascent with only a handful of studies. Sheng et al.~\cite{sheng2019unsupervised} introduce a dual-branch selection network: the visual stream processes target and reference frames, while the geometric stream handles depth and residual maps predicted by VO. Features are fused via cross-modal attention to compute a similarity score for keyframe selection. Messikommer et al.~\cite{messikommer2024reinforcement} use RL to train an agent that receives relative pose changes and keypoints as observations. These are processed using a multi-head attention mechanism followed by a two-layer MLP, and the agent outputs keyframe selection decisions. Dong et al.~\cite{dong2025kinnd} use real-time system status as input. Encode them with a four-layer MLP and extract features via hierarchical weighted self-attention. The probability of keyframe will be output by an MLP.

\section{FEED-FORWARD VISUAL ODOMETRY}
\begin{figure*}[ht!]
    \centering
    \includegraphics[width=1\linewidth]{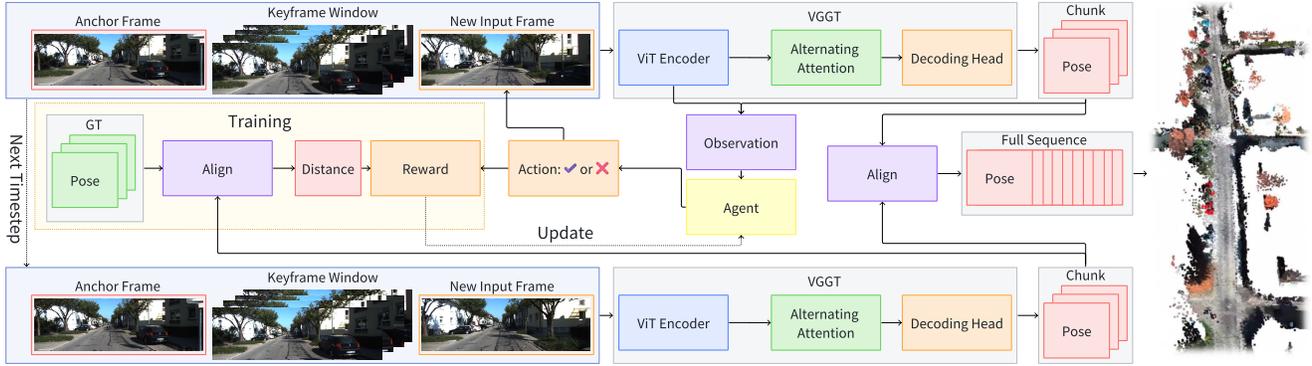}
    \vspace{-10pt}
    \caption{
    \textbf{Method Overview:} 
We propose a keyframe-based feed-forward VO with VGGT~\cite{wang2025vggt} as backbone and a RL-based keyframe method. The observation consists of the mean of CLS tokens extracted by the ViT encoder from all frames within the keyframe window, together with the relative pose changes. Based on this observation, the agent decides whether to retain the new input frame as a keyframe or not. After executing the action, the predicted poses are aligned with the ground-truth poses then be evaluated by the reward function, the result will be used to update the policy. During inference, the poses of all keyframe chunks are aligned by using the anchor frame, yielding the final pose estimation for the entire sequence. For clarity, some implementation details are omitted in the diagram to emphasize the overall framework and core workflow.
}
    \label{fig:main}
\end{figure*}

As illustrated in Fig.~\ref{fig:main}, our VO method takes the state-of-the-art visual foundation model VGGT~\cite{wang2025vggt} as the backbone. VGGT processes an unordered set of uncalibrated monocular RGB images as input. It first extracts per-image visual feature tokens using DINOv2~\cite{oquab2023dinov2}. Then alternates between frame-wise self-attention and global cross-attention layers. The resulting tokens are decoded through task-specific heads to produce essential outputs for downstream geometric tasks, including camera poses, depth maps, and point clouds.

\subsection{Sliding-Window}
Although VGGT's CUDA memory consumption scales sub-linearly with the number of input images, directly feeding a sequence of images into VGGT to obtain global poses for all frames remains computationally prohibitive for real-world visual localization tasks. A common strategy adopted in recent works, like VGGT-Long~\cite{deng2025vggt}, is to partition the input sequence into multiple chunks with a fixed number of overlapping frames between adjacent chunks. Each chunk is independently processed by VGGT to obtain local submap poses. The overlapping frames are subsequently utilized to align adjacent chunks via Sim(3) transformations, yielding a globally aligned trajectory. We adopt a conceptually similar philosophy to our method.

We maintain a fixed-size sliding window of the input frames. The first frame in the current window serves as the anchor frame, whose pose is already known in the global coordinate system. Unlike conventional overlap-based alignment between adjacent chunks, our approach leverages the anchor frame directly. All frames within the sliding window are fed into VGGT to obtain their relative poses to the anchor frame. These relative poses are then transformed into global poses using the known pose of the anchor frame.

After VGGT processing, the window content is summarized as an observation, which serves as the input to our RL agent. The agent decides whether the latest input frame should be retained as a keyframe.

\begin{itemize}
    \item \textbf{If the input frame is a keyframe}: 
      The sliding window shifts rightward. The current anchor frame is removed from the window, the second frame becomes the new anchor frame, and an empty position at the right end of the window is opened to accommodate the next input frame.
      
    \item \textbf{If the input frame is not a keyframe}: 
      It is immediately removed from the window. However, its relative pose change to the most recent keyframe is preserved.
\end{itemize}

\subsection{Initialization and Iterative Refinement}
\label{sec:pose-refinement}

We use 8 as the windows size for better training our RL agent within single RTX 4090. The first 7 input frames are directly treated as keyframe for fast initialization and we initialize the first frame of the entire sequence as the origin of the global coordinate system (i.e., identity pose). All subsequent poses are derived by accumulating the relative pose changes produced by VGGT within each sliding window, anchored to the current anchor frame.

This design offers two key advantages:
\begin{enumerate}
    \item Every frame has multiple opportunities to be refined whenever it appears in sliding windows with different anchor frames;
    \item The keyframe selection agent can actively control the spatial distribution of keyframes to achieve more uniform and high-quality coverage, ultimately leading to more accurate long-term trajectory estimation.
\end{enumerate}

\section{RL-BASED KEYFRAME METHOD}
Compared to conventional heuristic keyframe methods that rely on prior knowledge and require manual parameter tuning to adapt to different scenes, our approach aims to achieve adaptive, data-driven keyframe management.

\subsection{Problem Formulation}

We formulate this problem as a sequential decision-making task and address it using reinforcement learning. Specifically, the underlying foundation models and input images constitute the environment in which the RL agent operates. The agent, implemented as a neural network, jointly evaluates the existing keyframes within the current sliding window and the newly arrived frame, and determines whether to replace an existing keyframe with the new one or retain the current set. Each decision is followed by a reward signal that quantitatively evaluates the agent’s action, thereby guiding policy optimization and enabling the agent to make better decisions over time.

Based on this definition, we formulate the task as a Markov Decision Process (MDP), denoted as $M = \{S, A, P, R, \gamma\}$. $S$ represents the set of environment states. $A$ denotes the set of actions that can be executed to transition between states. $p(s'|s,a)$ is the probability of transitioning from current state $s \in S$ to next state $s' \in S$ after executing action $a \in A$.Each state transition is associated with a reward $r(s,a)$. Future rewards are discounted by $\gamma$. The agent is described by a stochastic policy $\pi(a \mid s)$, which specifies the probability of taking action $a \in A$ in state $s \in S$. Since the control policy is non-deterministic, it generates a probabilistic trajectory of states and actions, $\tau = \{s_{0}, a_{0}, s_{1}, a_{1}, \ldots, s_{t}, a_{t}\} $, abbreviated as $\tau \sim \pi$.

Following the control policy $\pi$, the expected sum of discounted rewards at each time step can be expressed by its value function:
\begin{equation}
V^{\pi}(s_i) = \mathbb{E}_{\tau \sim \pi} \Biggl[ \sum_{t=i}^{\infty} \gamma^{\,t-i} \, r(s_t, a_t) \Big|\ s_i \Biggr],
\end{equation}
The overall objective is to learn an optimal policy $ \pi^{*} $ that maximizes the expected cumulative discounted reward, where $ s_{0} $ denotes the initial state after the VO system is initialized:
\begin{equation}
\pi^{*} = \arg\max_{\pi} V^{\pi}(s_{0}).
\end{equation}

\subsection{Observation}
The core of our method is a trainable RL agent that selects an action $a \in \mathcal{A}$ based on a carefully designed observation $\mathbf{o}_t \in \mathcal{O}$, which approximates the underlying foundation models' internal state $s_t$. Given the inherent complexity of the foundation models' states, providing the agent with the complete state $s_t$ is neither computationally feasible nor beneficial for effective learning. Instead, we construct observation $\mathbf{o}_t$ as a selected subset of informative features derived from $s_t$. This design enables the policy $\pi: \mathcal{O} \to \Delta(\mathcal{A})$ to efficiently guide actions that enhance the performance of the foundation models. 

Transformer-based visual foundation models like VGGT, typically follows a similar processing pipeline. Specifically, high-dimensional feature representations are first extracted from the input images using vision foundation models such as DINOv2~\cite{oquab2023dinov2}. These features are then processed through multiple layers of self-attention and cross-attention to capture global context and cross-view relationships, before being decoded by task-specific heads to produce the desired outputs for various downstream tasks. Inspired by this architectural, we utilize the CLS token from each frame within the sliding windows, as output by DINOv2, mean-pooled it along with the relative pose of each frame to the new input frame, as the observation to the RL agent. The relative pose is normalized using the running mean std technique. On one hand, this design keeps the agent aligned with the actual input of VGGT, thereby avoiding information mismatch. On the other hand, the cls token provides a compact global semantic representation that preserves sensitivity to critical scene changes while significantly reducing the network size and computational overhead.

\subsection{RL Network and Action} 
The network architecture consists of a three-layer MLP with ReLU. The actor and critic are implemented as two separate networks sharing the same architecture.

The agent network outputs a discrete probability distribution over actions, represented by multiple independent categorical distributions. 
In our implementation, the agent evaluates the current sliding window with the new input frame, and subsequently chooses one of two possible actions above.

\subsection{Reward}
Another crucial component is the design of the reward function. At each timestep, the reward provides feedback that guides the agent towards desired behavioral patterns. In our task, the reward is computed immediately after the action is executed and before the next frame is processed by VGGT. The principal metric we use is the error between estimated and ground truth poses. This metric naturally addresses the sparse reward problem in RL due to its frequent availability in VO methods, making it the primary component of our reward function.

Specifically, We use the Umeyama algorithm to estimate the transformation that aligns the estimated trajectory with the ground truth using the first few poses in the sliding window. The translational error between the estimated and ground truth poses is then calculated and represented as $e_{tran}$. Since lower values of $e_{tran}$ are desired to obtain higher rewards, and $e_{tran}$ will never be 0, we introduce a threshold parameter $\lambda_{\mathrm{threshold}}$ to define an acceptable range of error. Specifically, the reward is computed based on $\lambda_{\mathrm{threshold}} - e_{tran}$, with a lower bound clipped at $-1$ to prevent excessively negative values~\cite{messikommer2024reinforcement}.

This approach reduces computational overhead for alignment while maintaining error within a reasonable range, preventing insufficient or excessive rewards due to variations in underlying methods.

We also introduce a compensation term $\alpha_{\mathrm{keyframe}}$, representing a small penalty or reward for adding a keyframe. This term is designed such that a significantly reduced translation error can outweigh it, thereby preventing action collapse that would otherwise lead to consistently removing or selecting keyframe. The reward function can be represented as 
\begin{equation}
\label{eq:reward}
\begin{split}
r(s_t, a_t) &= \lambda_1 \max\Bigl(-1, \lambda_{\mathrm{threshold}} - e_{\mathrm{tran}}(s_t, a_t)\Bigr) \\
&\quad + \lambda_2 \alpha_{\mathrm{keyframe}}(a_t)
\end{split}
\end{equation}
where $\lambda_{1} = 0.01$, $\lambda_{2} = 5\times10^{-3}$. The parameters $\lambda_{\mathrm{threshold}} = 0.2$ and $\alpha_{\mathrm{keyframe}} = 0.000025$.

\subsection{Training}
We utilize the PPO~\cite{schulman2017proximal} algorithm implemented in Stable Baselines3~\cite{raffin2021stable} to train our agent. The RL training process is divided into two alternating phases: the rollout phase and the policy update phase.

During the rollout phase, the agent collects information from the underlying method and the local map at each timestep to form $\mathbf{O_t}$. Based on these $\mathbf{O_t}$, the agent generates actions, which are then applied to the map and the underlying method to obtain rewards. All observations, actions, and rewards are collected, and this collected data is subsequently used in the policy update phase to improve the agent.

During training, the learning rate is linearly decayed from $3 \times 10^{-4}$ to $3 \times 10^{-5}$, and the Adam optimizer is employed.
 
We utilize the TartanAir synthetic dataset~\cite{wang2020tartanair} for training, which encompasses a wide range of simulated scenarios and includes challenging conditions such as low illumination and weather effects. 

We follow DPVO's approach by by applying random variations in brightness, contrast, and saturation during training, along with random frame dropping to enhance the agent's robustness and generalization.

To accelerate data collection during the rollout phase, we concurrently deployed 20 instances of the VGGT environment, each operating on distinct TartanAir sequences to gather training data. All uninitialized and error states were excluded from the training data to ensure data quality and improve training robustness.

Since performance varies across sequences, we adopt a privileged critic from ~\cite{messikommer2024reinforcement} to improve training stability. The privileged critic, with access to ground-truth poses of current and future states, can better discern whether tracking performance is affected by the agent’s actions or the intrinsic difficulty of the image sequence.

\section{EXPERIMENTS}
We trained our proposed method on the TartanAir~\cite{wang2020tartanair} synthetic dataset~\cite{wang2020tartanair}. For our comparative evaluation, we selected the real-world dataset: EuRoC~\cite{burri2016euroc}, TUM-RGBD~\cite{sturm2012benchmark} and KITTI~\cite{geiger2013vision}. All use the same model weights and evaluate trajectories based on the poses of all input frames.

To the best of our knowledge, VO methods  that leveraging visual foundation models, such as VGGT~\cite{wang2025vggt}, remain largely unexplored. Existing works primarily focus on SLAM method or in similar fashion including VGGT-SLAM~\cite{maggio2025vggt} and VGGT-Long~\cite{deng2025vggt}. Although directly comparing with VGGT-SLAM and VGGT-Long—both of which incorporate explicit post-processing, may not be entirely fair for our method, such comparisons nevertheless provide a valuable reference point for assessing the current performance and practical potential of the proposed method.

We also adopted VGGT-SLAM~\cite{maggio2025vggt} and VGGT-Long~\cite{deng2025vggt} with the same window size to ours and disabling their loop closure modules for fair comparison. For InfiniteVGGT~\cite{yuan2026infinitevggt}, we impose a 8GB CUDA memory constraint to ensure the entire sequence can be processed on a single RTX 4090 GPU. Regarding StreamVGGT~\cite{zhuo2025streaming} and FastVGGT~\cite{shen2025fastvggt}, which modify the original VGGT parameters, directly processing our evaluation sequences leads to out-of-memory (OOM) errors on a single RTX 4090. Consequently, we adopt the "chunk and align" strategy from VGGT-Long~\cite{deng2025vggt} for these methods and ensuring no further post-processing is performed.

Following the standard practices for evaluating VO methods, we have selected the Root Mean Square Error (RMSE) of Absolute Trajectory Error (ATE) as the metric for assessing performance. Before computing error metrics, the estimated trajectory is aligned with the ground-truth trajectory using the Umeyama algorithm.

\subsection{Comparison Analysis}

As shown in Tables~\ref{table:EuRoC},~\ref{table:TUM-RGBD}, and~\ref{table:KITTI}, the proposed method, trained exclusively on simulated datasets, achieves strong performance on real-world benchmarks, demonstrating notable generalization capability. Across the majority of sequences, our approach attains the best results and, in several cases, performs on par with or even surpasses methods that incorporate explicit post-processing modules. These results indicate that, for feed-forward VO frameworks, performance is not solely constrained by the lack of long-term memory; rather, effective preservation of current and short-term temporal context is equally critical. By explicitly maintaining keyframes, our method significantly enhances overall estimation accuracy and robustness.

Furthermore, the three evaluated datasets exhibit distinct characteristics. The experimental results reveal that methods augmented with explicit post-processing perform favorably in indoor environments with prominent geometric structures. However, their effectiveness degrades substantially in large-scale outdoor scenes with more complex and less constrained geometry, whereas the proposed approach maintains stable performance across diverse settings. This observation further highlights the potential of purely end-to-end models, a trend that is also evident from the strong performance of FastVGGT on the KITTI dataset.

\begin{table*}[!t]
\begin{center}
\caption{
Evaluation results on EuRoC. We report ATE [m] of the aligned trajectory for comparison.
The best and second-best results in each column are highlighted in red and orange, respectively.
}
\label{table:EuRoC}
\resizebox{\textwidth}{!}{
\begin{tabular}{l|cc|ccccccccccc|c}
\toprule
\textbf{Method} & 
\textbf{Keyframe} & \textbf{Post-Processing} & 
\textbf{MH01} & \textbf{MH02} & \textbf{MH03} & \textbf{MH04} & \textbf{MH05} &
\textbf{V101} & \textbf{V102} & \textbf{V103} &
\textbf{V201} & \textbf{V202} & \textbf{V203} & \textbf{Avg} \\
\midrule

VGGT-Long~\cite{deng2025vggt} & \textcolor{red}{\ding{55}} & \textcolor{red}{Required}  & \colorbox{red!50}{2.11} & \colorbox{red!50}{2.19} & {3.10} & \colorbox{red!50}{4.65} & \colorbox{orange!50}{4.99} & \colorbox{orange!50}{1.46} & \colorbox{orange!50}{1.60} & {1.48} & \colorbox{orange!50}{1.68} & \colorbox{orange!50}{1.79} & {1.91} & \colorbox{orange!50}{2.45} \\
VGGT-SLAM~\cite{maggio2025vggt} & \textcolor{green}{\ding{51}} & \textcolor{red}{Required}  & 4.78 & 4.91 & 3.39 & 6.79 & 5.41 & 1.81 & {1.68} & \colorbox{red!50}{1.37} & 2.36 & 1.91 & 1.96 & 3.31 \\
\midrule
FastVGGT~\cite{shen2025fastvggt}   &  \textcolor{red}{\ding{55}} & \textcolor{green}{No}  & 4.58 & 4.92 & 4.42 & 6.71 & 4.93 & 2.28 & 2.36 & 1.81 & 2.32 & 2.54 & 2.52 & 3.58 \\
StreamVGGT~\cite{zhuo2025streaming}  & \textcolor{red}{\ding{55}} & \textcolor{green}{No} & 4.29 & 4.62 & 3.56 & 6.80 & 6.85 & 1.84 & 1.76 & 1.57 & 2.28 & 2.09 & 1.92 & 3.42 \\
InfiniteVGGT~\cite{yuan2026infinitevggt}  &\textcolor{red}{\ding{55}} & \textcolor{green}{No}  & {3.48} & 4.40 & \colorbox{orange!50}{2.77} & 5.45 & 6.34 & {1.63} & \colorbox{red!50}{1.57} & \colorbox{orange!50}{1.39} & {1.98} & {1.85} & \colorbox{red!50}{1.78} & {2.97} \\
\midrule
Ours  & \textcolor{green}{\ding{51}}  & \textcolor{green}{No}    & \colorbox{orange!50}{2.74} & \colorbox{orange!50}{2.90} & \colorbox{red!50}{2.63} & \colorbox{orange!50}{4.82} & \colorbox{red!50}{4.68} & \colorbox{red!50}{1.30} & 1.69 & 1.50 & \colorbox{red!50}{0.99} & \colorbox{red!50}{1.71} & \colorbox{orange!50}{1.86} & \colorbox{red!50}{2.44} \\
\bottomrule
 \end{tabular}}
 \vspace{-5pt}

\end{center}
\end{table*}

\begin{figure}[!t]
\centering
\includegraphics[width=0.99\linewidth]{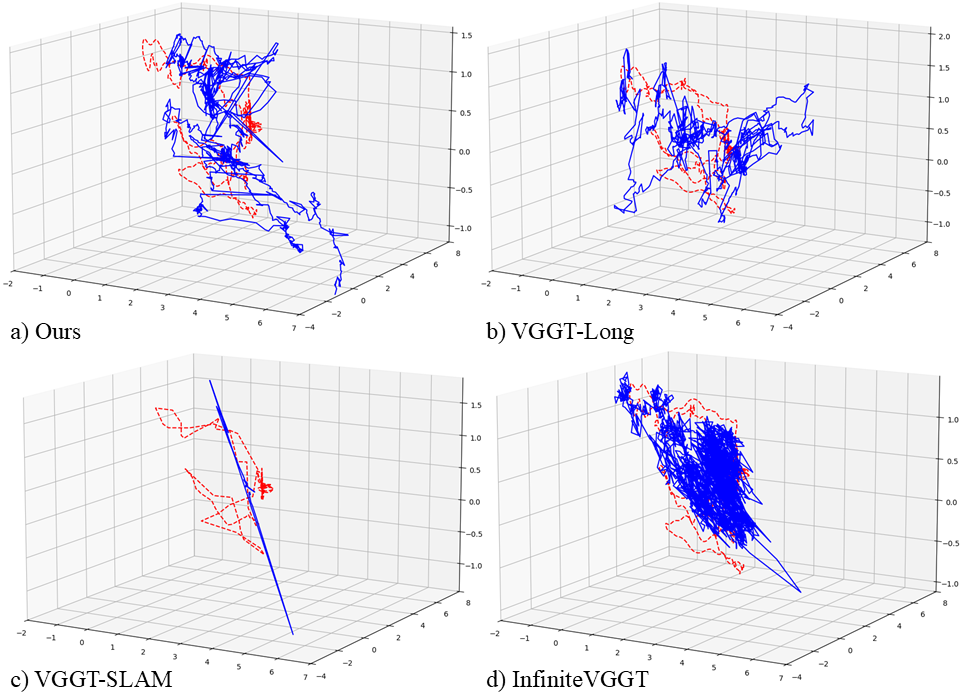}
\vspace{-5pt}
\caption{Visualized trajectory of our method, VGGT-Long, VGGT-SLAM and InfiniteVGGT on MH\_02 from EuRoC. The ground-truth trajectory is shown in red, while the estimated trajectories are shown in blue.}
\label{fig:traj_compare1}
\end{figure}

\begin{figure}[!t]
\centering
\includegraphics[width=0.99\linewidth]{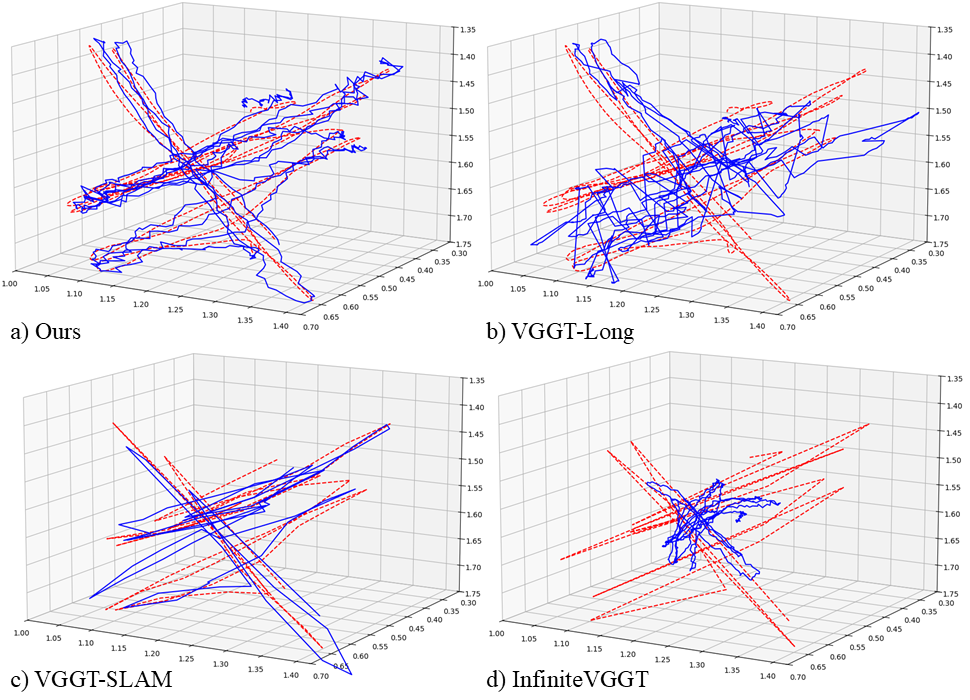}
\vspace{-5pt}
\caption{Visualized trajectory of our method, VGGT-Long, VGGT-SLAM and InfiniteVGGT on fr1-xyz from TUM-RGBD. The ground-truth trajectory is shown in red, while the estimated trajectories are shown in blue.}
\label{fig:traj_compare2}
\end{figure}

\begin{figure}[!t]
\centering
\includegraphics[width=0.99\linewidth]{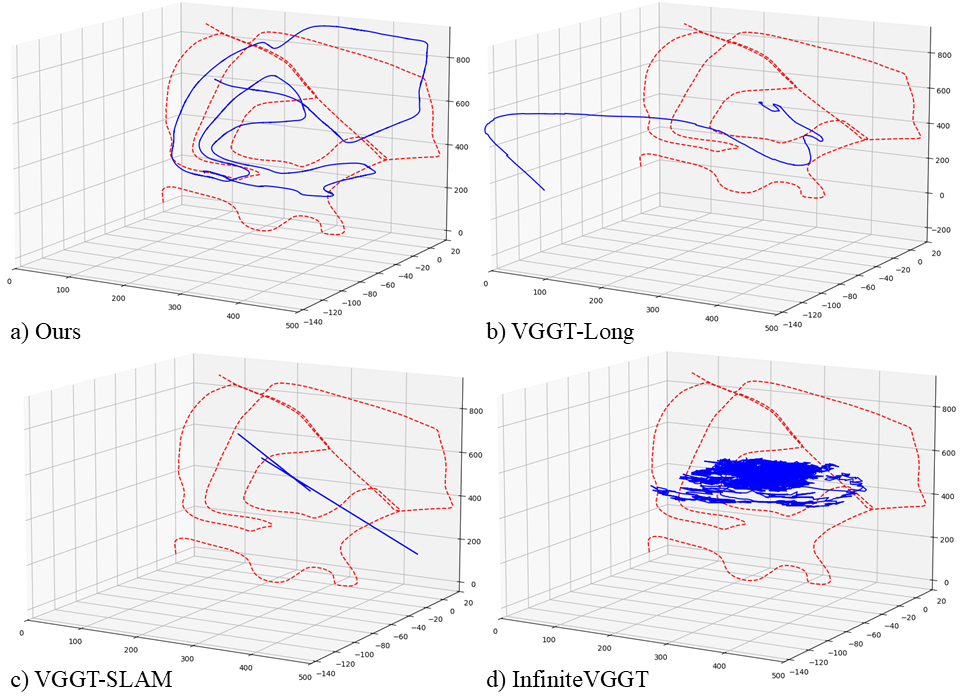}
\vspace{-5pt}
\caption{Visualized trajectory of our method, VGGT-Long, VGGT-SLAM and InfiniteVGGT on 02 from KITTI. The ground-truth trajectory is shown in red, while the estimated trajectories are shown in blue.}
\label{fig:traj_compare3}
\end{figure}

\begin{table*}[!t]
\centering
\caption{
Evaluation results on TUM-RGBD. We report ATE [m] of the aligned trajectory for comparison.
The best and second-best results in each column are highlighted in red and orange, respectively.
}
\label{table:TUM-RGBD}
\resizebox{\textwidth}{!}{
\begin{tabular}{l|cc|ccccccccc|c}
\toprule
\textbf{Method} & \textbf{Keyframe} & \textbf{Post-Processing} 
    & \textbf{fr1-360} 
    & \textbf{fr1-desk} 
    & \textbf{fr1-desk2} 
    & \textbf{fr1-floor} 
    & \textbf{fr1-plant} 
    & \textbf{fr1-room} 
    & \textbf{fr1-rpy} 
    & \textbf{fr1-teddy} 
    & \textbf{fr1-xyz} 
    & \textbf{Avg} \\
\midrule
VGGT-Long~\cite{deng2025vggt}& \textcolor{red}{\ding{55}} & \textcolor{red}{Required} & \colorbox{orange!50}{0.142} & \colorbox{red!50}{0.138} & \colorbox{red!50}{0.155} & \colorbox{red!50}{0.253} & \colorbox{orange!50}{0.197} & \colorbox{red!50}{0.283} & 0.057 & \colorbox{orange!50}{0.208} & 0.087 & \colorbox{red!50}{0.169} \\
VGGT-SLAM~\cite{maggio2025vggt}  & \textcolor{green}{\ding{51}} & \textcolor{red}{Required} & \colorbox{red!50}{0.130} & 0.216 & 0.568 & 0.752 & \colorbox{red!50}{0.076} & 0.770 & \colorbox{orange!50}{0.048} & 0.224 & \colorbox{orange!50}{0.042} & 0.314 \\
\midrule
FastVGGT~\cite{shen2025fastvggt} & \textcolor{red}{\ding{55}} & \textcolor{green}{No} & 0.178 & 0.433 & 0.456 & 0.695 & 0.862 & 1.109 & 0.047 & 0.794 & 0.171 & 0.527 \\
StreamVGGT~\cite{zhuo2025streaming} & \textcolor{red}{\ding{55}} & \textcolor{green}{No} & 0.207 & 0.869 & 0.963 & 0.789 & 0.751 & 1.014 & 0.063 & 0.987 & 0.185 & 0.648 \\
InfiniteVGGT~\cite{yuan2026infinitevggt} & \textcolor{red}{\ding{55}} & \textcolor{green}{No} & 0.191 & 0.496 & 0.546 & 0.699 & 0.430 & 0.920 & 0.053 & 0.504 & 0.177 & 0.446 \\
\midrule
Ours & \textcolor{green}{\ding{51}}  & \textcolor{green}{No}  & 0.147 & \colorbox{orange!50}{0.176} & \colorbox{orange!50}{0.163} & \colorbox{orange!50}{0.345} & 0.240 & \colorbox{orange!50}{0.409} & \colorbox{red!50}{0.037} & \colorbox{red!50}{0.144} & \colorbox{red!50}{0.015} & \colorbox{orange!50}{0.186} \\
\bottomrule
\end{tabular}}
\vspace{-5pt}

\end{table*}

\begin{table*}[!t]
\centering
\caption{
Evaluation results on KITTI. We report ATE [m] of the aligned trajectory for comparison.
The best and second-best results in each column are highlighted in red and orange, respectively.
}
\label{table:KITTI}
\resizebox{\textwidth}{!}{
\begin{tabular}{l|cc|ccccccccccc|c}
\toprule
\textbf{Method}  & \textbf{Keyframe} & \textbf{Post-Processing} 
    & \textbf{00} 
    & \textbf{01} 
    & \textbf{02} 
    & \textbf{03} 
    & \textbf{04} 
    & \textbf{05} 
    & \textbf{06} 
    & \textbf{07} 
    & \textbf{08} 
    & \textbf{09} 
    & \textbf{10} 
    & \textbf{Avg} \\
\midrule
VGGT-Long~\cite{deng2025vggt} & \textcolor{red}{\ding{55}} & \textcolor{red}{Required}  & {181.9} & {232.2} & {262.8} & 30.5 & 23.8 & {147.2} & \colorbox{red!50}{71.8} & \colorbox{orange!50}{58.6} & {247.7} & {159.5} & {71.2} & {135.2} \\
VGGT-SLAM~\cite{maggio2025vggt} & \textcolor{green}{\ding{51}} & \textcolor{red}{Required} & 190.8 & 793.3 & 303.8 & 167.8 & \colorbox{orange!50}{4.9} & 167.7 & 149.9 & 90.2 & 271.51 & 218.1 & 221.2 & 234.5 \\
\midrule
FastVGGT~\cite{shen2025fastvggt} & \textcolor{red}{\ding{55}} & \textcolor{green}{No} & \colorbox{red!50}{102.5} & \colorbox{red!50}{176.8} & \colorbox{orange!50}{170.0} & \colorbox{red!50}{11.2} & \colorbox{red!50}{2.5} & \colorbox{red!50}{76.1} & 131.2 & 61.5 & \colorbox{orange!50}{99.3} & \colorbox{orange!50}{99.3} & \colorbox{orange!50}{36.7} & \colorbox{orange!50}{87.9} \\
StreamVGGT~\cite{zhuo2025streaming} & \textcolor{red}{\ding{55}} & \textcolor{green}{No} & 193.6 & 716.8 & 305.4 & 164.8 & 100.4 & 160.0 & 135.3 & 90.6 & 264.8 & 224.0 & 210.9 & 233.3\\
InfiniteVGGT~\cite{yuan2026infinitevggt} & \textcolor{red}{\ding{55}} & \textcolor{green}{No} & 181.0 & 607.9 & 296.1 & 157.1 & 72.5 & 157.0 & 121.2 & 76.2 & 261.6 & 198.6 & 157.3 & 207.9 \\
\midrule
Ours & \textcolor{green}{\ding{51}}  & \textcolor{green}{No} & \colorbox{orange!50}{138.1} & \colorbox{orange!50}{179.1} & \colorbox{red!50}{153.5} & \colorbox{orange!50}{12.0} & 13.2 & \colorbox{orange!50}{131.6} & \colorbox{orange!50}{73.0} & \colorbox{red!50}{45.9} & \colorbox{red!50}{86.5} & \colorbox{red!50}{97.5} & \colorbox{red!50}{26.8} & \colorbox{red!50}{87.0} \\
\bottomrule
\end{tabular}
}
\vspace{-5pt}

\end{table*}

\subsection{Comparison of Different Keyframe Strategies}

The previous experiment demonstrates the effectiveness of the proposed approach and highlights the importance of keyframe selection for feed-forward VO frameworks. In this experiment, we further investigate the impact of different keyframe decision strategies on overall performance. To the best of our knowledge, most existing VGGT-based visual localization pipelines do not perform explicit input frame selection.

Accordingly, we adopt VGGT-SW and VGGT-LK as baseline methods. VGGT-SW employs the same sliding window mechanism described above but treats every incoming frame as a keyframe. Inspired by VGGT-SLAM, VGGT-LK also uses an identical sliding window, while incorporating Lucas–Kanade (LK) optical flow with a fixed threshold of 50 to determine keyframes. Aside from the keyframe strategy, both baselines share the same pipeline as the proposed method, and no additional post-processing is applied.

\begin{figure}[!t]
\centering
\includegraphics[width=0.99\linewidth]{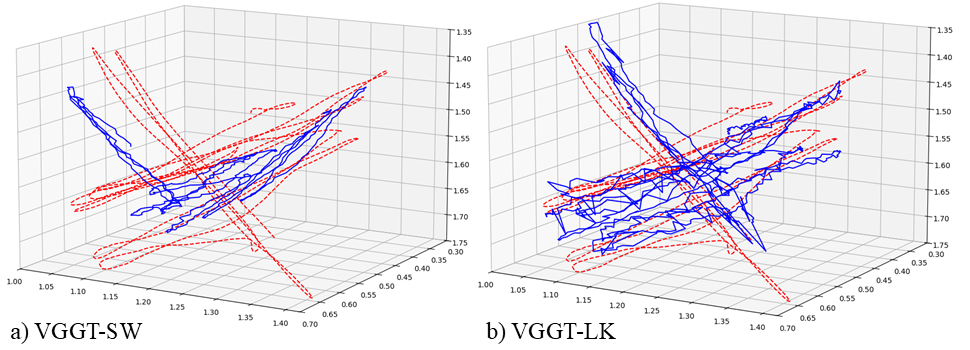}
\vspace{-5pt}
\caption{Visualized trajectory of VGGT-SW and VGGT-LK on fr1-xyz from TUM-RGBD. The ground-truth trajectory is shown in red, while the estimated trajectories are shown in blue.}
\label{fig:traj_our_lk_compare}
\end{figure}

As shown in Table~\ref{table:keyframe} and visualized in Fig.~\ref{fig:traj_our_lk_compare}, the proposed method consistently reduces the average ATE across KITTI, TUM-RGBD, and EuRoC, three datasets with markedly different characteristics. These results suggest that, for VGGT-based feed-forward frameworks, directly transferring conventional geometry-driven keyframe selection heuristics does not align well with the intrinsic properties of visual foundation models. Consequently, VGGT-SW and VGGT-LK exhibit varying performance across different scenes. In contrast, the proposed data-driven keyframe policy is better matched to the characteristics of VGGT, leading to more robust and consistently superior performance across all evaluated datasets.

\begin{table}[!t]
\centering
\caption{Evaluation of VGGT with different keyframe selection methods on TUM-RGBD and EuRoC datasets. Average ATE (m) is reported.}
\label{table:keyframe}
\resizebox{0.65\linewidth}{!}{
\begin{tabular}{l|c|c|c}
\toprule
\textbf{Method} & \multicolumn{2}{c}{\textbf{Avg. ATE (m)}} \\
\cmidrule(r){2-4}
& \textbf{KITTI} &\textbf{TUM-RGBD} & \textbf{EuRoC} \\
\midrule
VGGT-SW & 88.3  & 0.233 & 2.64 \\
VGGT-LK & 109.9 & 0.194 & 2.54 \\
\midrule
\textbf{Ours} & \textbf{87.0} & \textbf{0.186} & \textbf{2.44} \\
\bottomrule
\end{tabular}
}

\end{table}

\subsection{Runtime and Ablations}
As reported in Table~\ref{table:runtime}, the proposed RL-based keyframe method introduces negligible computational overhead to the overall system.

\begin{table}[!t]
\centering
\caption{Runtime breakdown of major components of our method on the EuRoC dataset. We report the average execution time (ms) over full-dataset runs.}
\label{table:runtime}
\resizebox{0.6\linewidth}{!}{
\begin{tabular}{l|c}
\toprule
\textbf{Component} & \textbf{Avg. Time (ms)} \\
\midrule
\textbf{Ours (Total)}        & 380.2 \\
\midrule
Aggregator                  & 370.0 \\
DPT                          & 1.1 \\
\textbf{Keyframe Decision}  & \textbf{0.74} \\
\bottomrule
\end{tabular}
}

\end{table}

\begin{table}[!t]
\centering
\caption{Ablation study of the proposed method on the EuRoC dataset. Average ATE is reported.}
\label{table:Ablations}
\resizebox{0.45\linewidth}{!}{
\begin{tabular}{l|c}
\toprule
\textbf{Method} & \textbf{Avg. ATE(m)} \\
\midrule
VGGT-SW & 2.648 \\
\midrule
\textbf{Full Method} & \textbf{2.443} \\
w/o $o_{\text{pose}}$ & 2.661 \\
w/o $o_{\text{cls\_token}}$ & 2.647 \\
w/o $\alpha_{\text{keyframe}}$ & 2.648\\
\bottomrule
\end{tabular}
}

\end{table}

As shown in Table~\ref{table:Ablations}, the full version of our method achieves the lowest average ATE on the EuRoC dataset, confirming the effectiveness of the proposed method.
Removing any information of the observation will cause performance drop. And the $\alpha_{\text{keyframe}}$ plays an important role for prevent action collapse.

Overall, these ablation results demonstrate that every component of proposed method are necessary to achieve robust and accurate VO.

\section{CONCLUSIONS}

In this paper, we propose a keyframe-based feed-forward VO method built upon the visual foundation model VGGT.
By incorporating an RL-based keyframe method, our method effectively enhances the localization capability of VGGT without modifying its core architecture.
The proposed keyframe method is a fully data-driven approach that leverages reinforcement learning to make adaptive and informed keyframe decisions, rather than relying on hand-crafted geometric heuristics.
Extensive experimental results demonstrate that our VO method generalizes well across diverse scenes and datasets, only trained in simulated datasets.
We believe that this work highlights the importance of keyframe-based memory maintenance for feed-forward VO and provides a promising direction for future advancements toward more robust and scalable visual odometry systems. 

In future work, we will investigate mechanisms for maintaining long-term memory(loop closing), with the goal of extending the proposed method toward a fully feed-forward SLAM framework.








\bibliographystyle{IEEEtran}
\bibliography{ref}

\end{document}